\documentclass[lettersize,journal]{IEEEtran}
\usepackage{amsmath,amsfonts}
\usepackage{array}
\usepackage[caption=false,font=normalsize,labelfont=sf,textfont=sf]{subfig}
\usepackage{textcomp}
\usepackage{stfloats}
\usepackage{url}
\usepackage{verbatim}
\usepackage{graphicx}
\usepackage{cite}
\usepackage{diagbox}
\usepackage{makecell}

\usepackage{amsmath,amssymb,amsfonts}
\usepackage{algpseudocode}
\usepackage{booktabs}

\usepackage{xcolor}
\usepackage{url} 
\usepackage{subfig}
\usepackage[ruled]{algorithm2e}
\usepackage{multirow}

\newtheorem{myDef}{Definition}
\begin{document}

\title{Transaction Fraud Detection via an Adaptive Graph Neural Network}

\author{Yue~Tian,
        Guanjun~Liu,~\IEEEmembership{Senior Member,~IEEE,}
	Jiacun~Wang,~\IEEEmembership{Senior Member,~IEEE,}
	and~Mengchu~Zhou,~\IEEEmembership{Fellow,~IEEE}
\thanks{This work has been submitted to the IEEE for possible publication. Copyright may be transferred without notice, after which this version may no longer be accessible.}
\thanks{Yue Tian and Guanjun Liu are with Department of Computer Science, Tongji University, Shanghai 201804, China (e-mail: 1810861@tongji.edu.cn; liuguanjun@tongji.edu.cn).}
\thanks{Jiacun Wang is with the Department of Computer Science and Software Engineering, Monmouth University, W. Long Branch, NJ 07764, USA (e-mail: jwang@monmouth.edu).}
\thanks{Mengchu Zhou is with the Department of Electrical and Computer Engineering, New Jersey Institute of Technology, Newark, NJ 07102 USA (e-mail: zhou@njit.edu).}

}

\markboth{Journal of \LaTeX\ Class Files,~Vol.~14, No.~8, August~2021}%
{Shell \MakeLowercase{\textit{et al.}}: A Sample Article Using IEEEtran.cls for IEEE Journals}


\maketitle

\begin{abstract}
Many machine learning methods have been proposed to achieve accurate transaction fraud detection, which is essential to the financial security of individuals and banks. However, most existing methods leverage original features only or require manual feature engineering. They lack the ability to learn discriminative representations from transaction data. Moreover, criminals often commit fraud by imitating cardholders' behaviors, which causes the poor performance of existing detection models. In this paper, we propose an Adaptive Sampling and Aggregation-based Graph Neural Network (ASA-GNN) that learns discriminative representations to improve the performance of transaction fraud detection. A neighbor sampling strategy is performed to filter noisy nodes and supplement information for fraudulent nodes. Specifically, we leverage cosine similarity and edge weights to adaptively select neighbors with similar behavior patterns for target nodes and then find multi-hop neighbors for fraudulent nodes. A neighbor diversity metric is designed by calculating the entropy among neighbors to tackle the camouflage issue of fraudsters and explicitly alleviate the over-smoothing phenomena. Extensive experiments on three real financial datasets demonstrate that the proposed method ASA-GNN outperforms state-of-the-art ones.
\end{abstract}


\section{Introduction}
\IEEEPARstart{O}{nline} transaction is a popular and convenient way of electronic payment. It also increases the incidences of financial fraud and causes massive monetary losses to individuals and banks. The global losses reached 25 billion dollars in 2018 and have kept increasing \cite{ref1}. According to statistics from Nilson Report, the losses jumped to 28.65 billion in 2020 \cite{ref2}. Financial institutions have taken measures to prevent fraud. In traditional methods, online transactions are checked against some expert rules, and then suspicious transactions are fed to a detection model. The task of the detection model is to mine fraud patterns (represented as some rules) from sizeable historical transaction data so that the model can find transactions that match these rules. However, the ability of these rules is limited and hardly adapts to the fast changes in fraud patterns.

How to quickly mine and represent as many fraud patterns as possible and fit their changes is complicated since fraudsters and detectors of fraud transactions have kept a dynamic gaming process for a long time \cite{ref3}. Transaction records often contain transaction-related elements, such as location, date, time, and relations. Although there are machine learning-based methods to detect fraudulent transactions, most require manual feature engineering based on the above elements and the construction of supervised classifiers  \cite{ref5,ref6}. These methods fail to automatically detect fraud patterns and express important behavior information \cite{ref7}. On the one hand, there are many interactions among transactions \cite{ref9}. On the other hand, the transaction behaviors of users are dynamic \cite{ref10}. Hence, designing a more discriminative representation framework for transaction fraud detection remains a big challenge. The camouflage of fraudsters is another challenge that causes performance degradation and poor generalization of many detection approaches \cite{ref11}.

Recently, graph neural networks (GNNs) have been used to learn representations automatically for some prediction tasks \cite{ref12,ref51,ref52}.
In contrast to traditional machine learning methods, GNN utilizes a neighborhood aggregation strategy to learn representations and then uses a neural network for node classification and link prediction \cite{ref15}. These methods can capture rich interactions among samples and avoid feature engineering \cite{ref11}. However, learning discriminative representations by directly applying these graph techniques to our transaction fraud detection problem is challenging. They ignore the relationship among features and the dynamic changes in cardholders’ behaviors. We visualize the representations of the general GNN models, including GraphSAGE and GCN. As shown in Figs.~1 and 2, they fail to distance the fraudulent transactions from the legitimate ones. Moreover, GNNs face an over-smoothing problem (indistinguishable representations of nodes in different classes), which results from the over-mixing of information and noise \cite{ref16,ref56,Pourhabibi}. In the applications of transaction fraud detection, the fact that fraudulent nodes are connected to legitimate ones by disguising the cardholders' behaviors \cite{ref11}, as shown in Fig.~3, exacerbates the effects of the over-smoothing issue.

\begin{figure}
  \centering
  \subfloat[]{\includegraphics[width=1.65in]{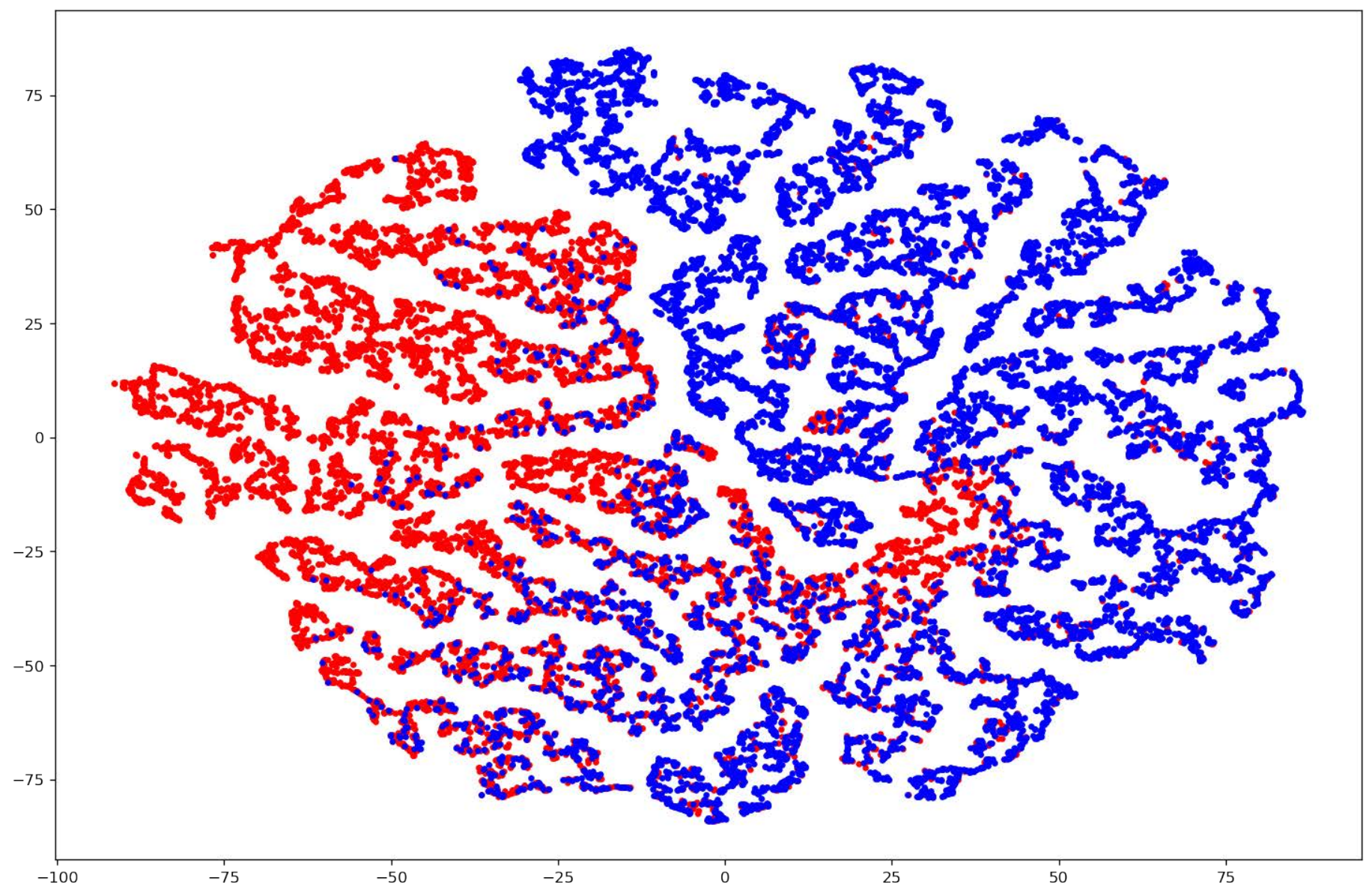}}
  \subfloat[]{\includegraphics[width=1.65in]{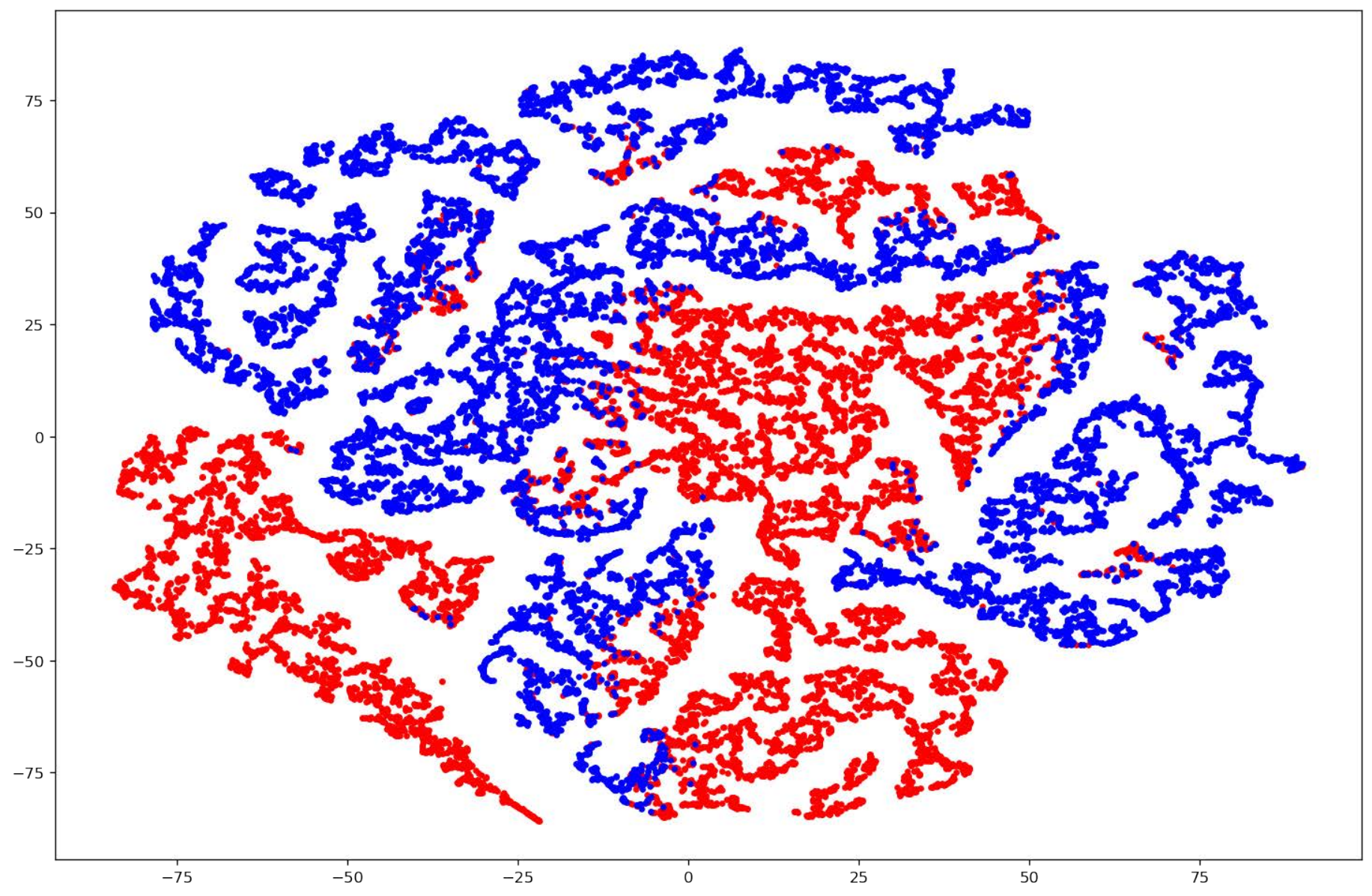}}
  \caption{GraphSAGE's visualization results. The red nodes represent fraudulent transactions, while the others represent legitimate ones. (a) Before training on our financial dataset. (b) After its training on our financial dataset.}
\end{figure}

\begin{figure}
  \centering
  \subfloat[]{\includegraphics[width=1.65in]{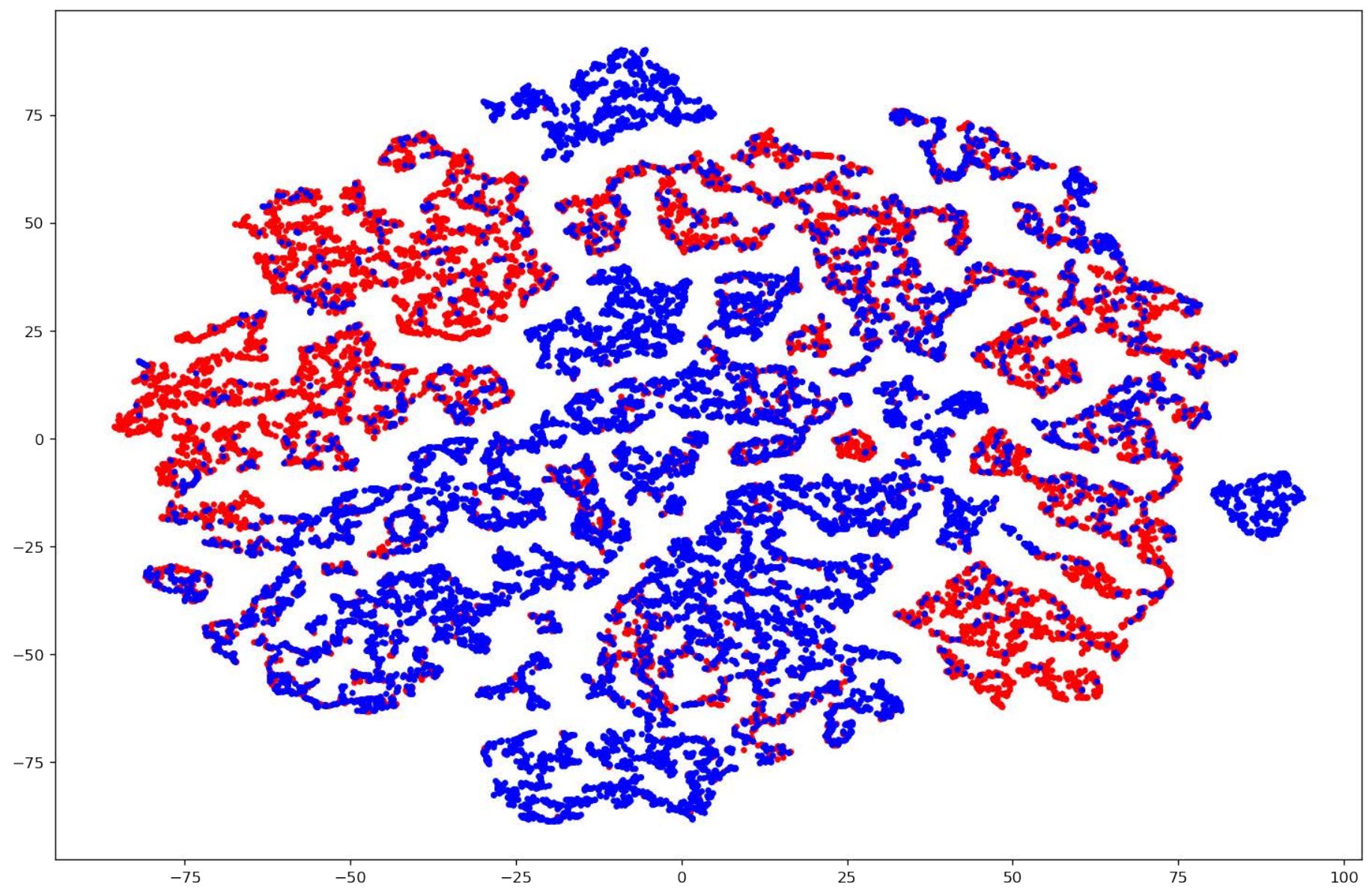}}
  \subfloat[]{\includegraphics[width=1.65in]{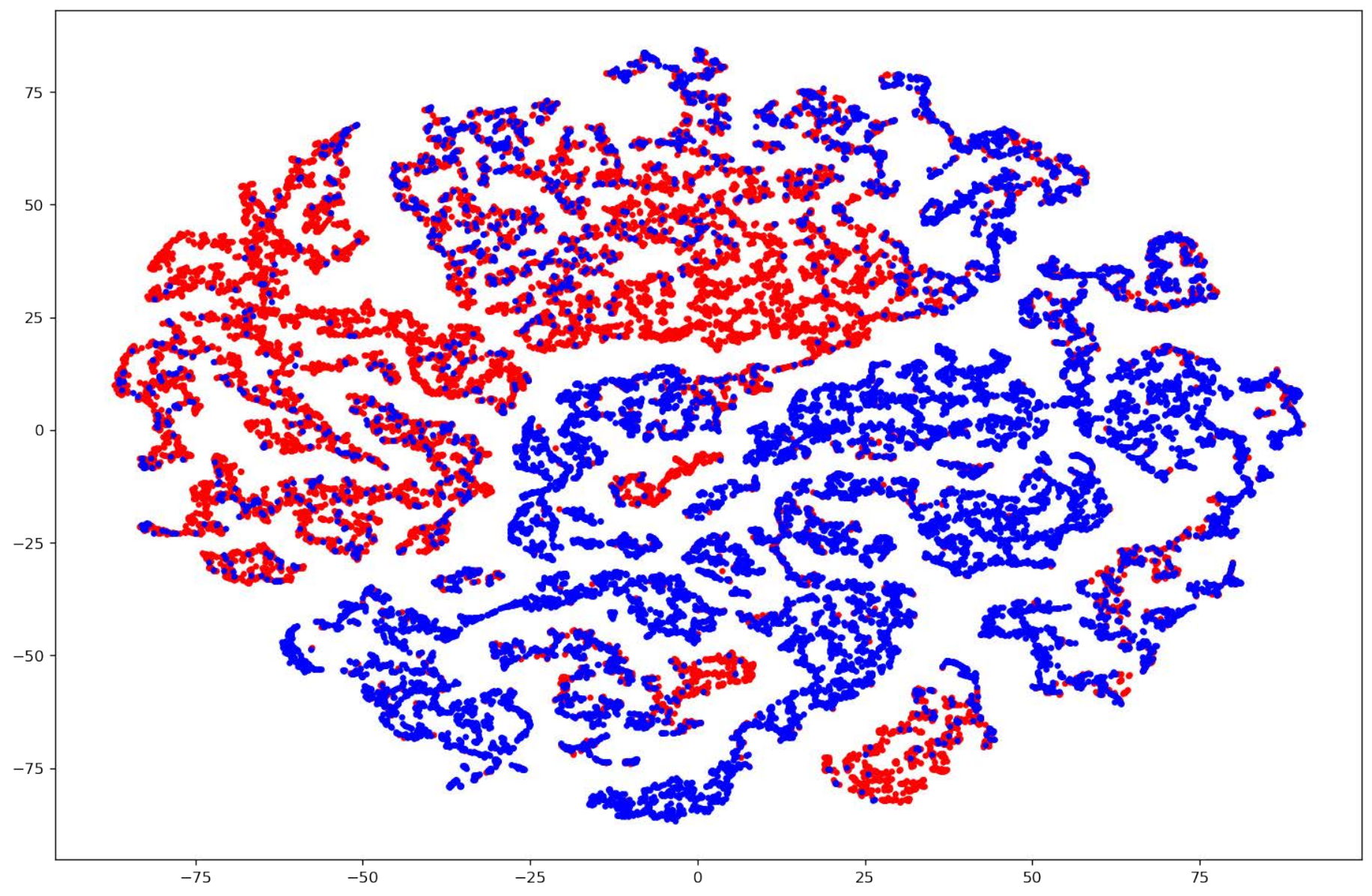}}
  \caption{GCN's visualization results.The red nodes represent fraudulent transactions, while the others represent legitimate ones. (a) Before training on our financial dataset. (b) After its training on our financial dataset.}
\end{figure}

To tackle the above problem, we propose an Adaptive Sampling and Aggregation-based GNN for transaction fraud detection, named ASA-GNN. It integrates our newly proposed Adaptive Sampling and Aggregation methods. First, we use raw transaction records to construct the transaction graph, which considers the relationship of features and the dynamic changes in cardholders' behaviors. Based on it, we design a sampler to filter as many noisy neighbors as possible while retaining structural information. Cosine similarity and edge weight are used to select similar neighbor nodes. Then, we over-sample these neighbor nodes to tackle the need for more links among fraudulent nodes. To deal with the camouflage issue of fraudsters, a neighbor diversity metric is defined and calculated based on the entropy among neighbor nodes to distinguish whether neighborhood aggregation is harmful. Each node has its neighborhood aggregation degree. As a result, intraclass compactness and interclass separation can be guaranteed. This work aims to make the following new contributions:
\begin{enumerate}[]
    \item We propose a graph neural network that learns discriminative representations to improve the performance of transaction fraud detection.
     \item We propose a new sampling strategy to filter noisy nodes and capture neighbors with the same behavior pattern for fraudulent nodes based on the distance between two nodes measured by cosine similarity and edge weight.
     \item We define a neighbor diversity metric to make each node adaptive in its aggregation process, which handles the camouflage issue of fraudsters and alleviates the over-smoothing phenomena. 
    \item Extensive experiments conducted on three financial datasets show that the proposed ASA-GNN achieves significant performance improvements over traditional and state-of-the-art methods. 

\end{enumerate}

The rest of this paper is organized as follows. Section II presents the related work. Section III describes the proposed ASA-GNN. Section IV presents three real datasets and discusses the experimental results of performance comparison, ablation studies, and parameter sensitivity analysis. Section V concludes the paper.

\begin{figure}[!t]
	\centering
	\includegraphics[width=3.5in]{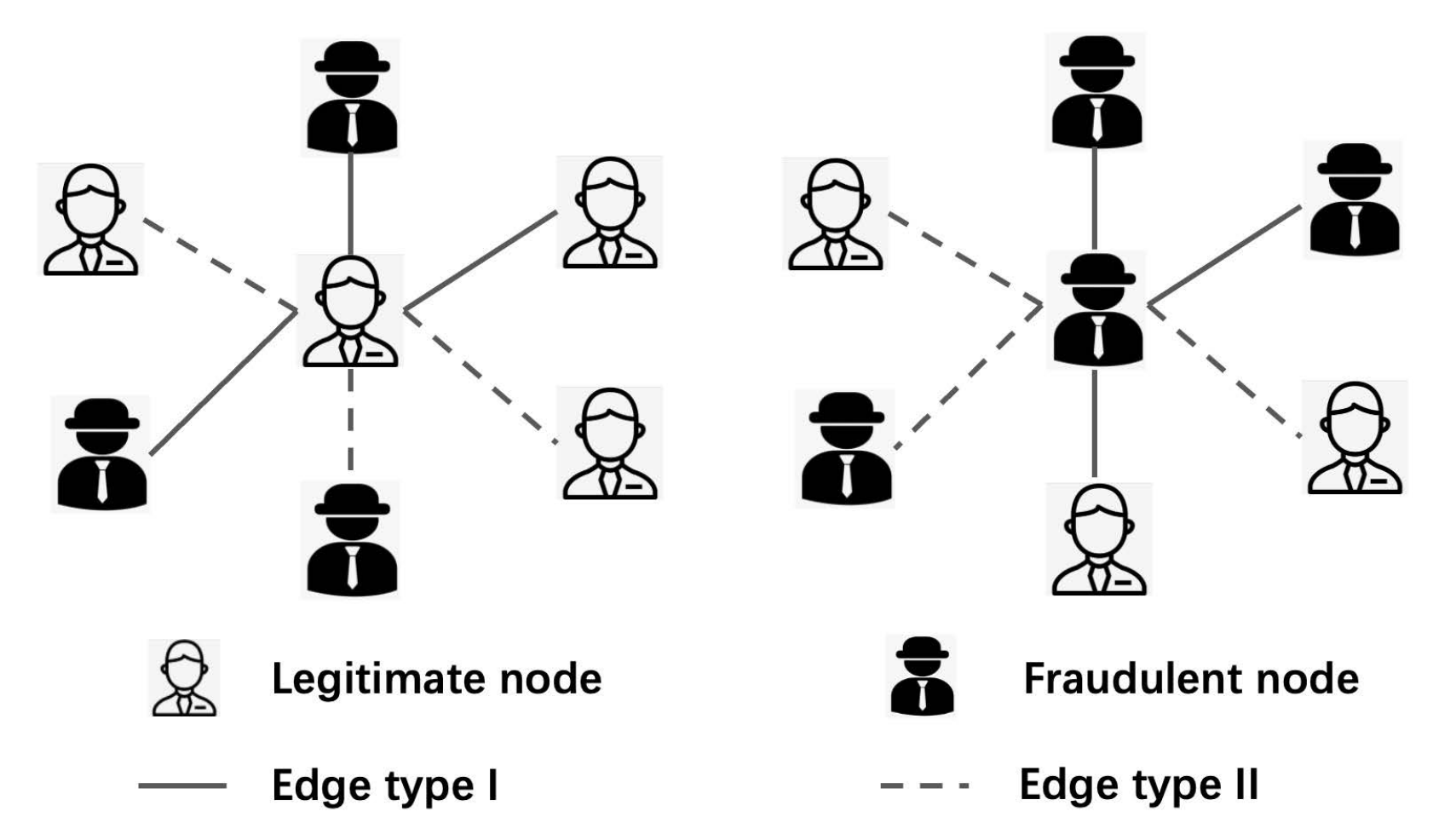}
	\caption{The camouflage issue of fraudsters. Fraudsters attenuate their suspicions by disguising the cardholders' behaviors such that the detection system thinks they are legitimate transactions.}
	\label{fig:1.1}       
\end{figure}

\section{Background and Related Work}

\subsection{Transaction Fraud Detection Model}
Researchers have proposed many methods based on expert rules and machine learning in many fields including transaction fraud detection tasks, which have achieved much success \cite{Dornadula, Jin, Lee, Jinhu}. Their core is to learn some information from historical data to detect fraudulent transactions automatically. They have been proven effective for known fraud patterns but cannot deal with unknown fraud types \cite{ref42}. Experts have started to use deep learning methods to solve it. Therefore, according to the correlations among the transaction features, deep learning methods can automatically capture cross-feature relationships so that these transaction records can be accurately portrayed, which helps detect fraud behaviors \cite{ref43}. The Convolutional Neural Network (CNN) method is one of the commonly used methods \cite{ref43}.

In addition to the relationship among transaction features, existing feature engineering methods extract the association of transaction records to improve performance \cite{ref8}. The aggregation strategy is a classical feature engineering method for transaction fraud detection. It groups the transaction records according to the specified time and then extracts the amount-related features and numbers of these records as the aggregation features \cite{ref48}. Location and merchant code are also considered and used to generate aggregation features, increasing the user's periodic behavior information \cite{ref7}. Moreover, some methods, such as a Recurrent Neural Network (RNN) method \cite{ref57}, start to explore the dynamic information of transactions versus time \cite{ref47,ref10}. A Long Short-Term Memory (LSTM) network relies on the evolution of data distribution versus time to capture the dynamic information \cite{ref47}. By considering the small and various changes in data distribution, an ensemble method is proposed to achieve a better performance \cite{ref10}.

To comprehensively focus on the various mentioned relationships, researchers have utilized transaction records to construct a graph \cite{ref49}. For example, GNN can capture the relationships among transactions and achieve better performance in fraud detection. However, since this method uses only one feature to construct a sparse graph, it fails to mine many useful fraud features \cite{ref49}.

Most of the mentioned approaches fail to comprehensively consider the relationship among transactions,  the relationship among the transaction features, and dynamic change information of cardholders' behaviors. In our previous work \cite{ref53}, we constructed a weighted multigraph to tackle the challenge since it can use multiple features as long as logic propositions can represent them. 
Based on this weighted multigraph, we use GNN to extract the above relationships and the dynamic changes. However, the method has some shortcomings, as stated in Section I. This work is motivated by the need to overcome such shortcomings.

\subsection{GNN}

A GNN is an effective framework that can learn graph representations by modeling the relationships of non-Euclidean graph data \cite{ref15}. The concept is initially outlined in \cite{ref19}, where a convolution operation in image processing is used in the graph data processing. After that, several graph-based methods are proposed and applied. Most earlier algorithms obtain the embedding representations in two steps: 1) Obtain a sequence of neighbor nodes for each node by using a random walk strategy; 2) Use machine learning models to gain the topological structure information of a graph and obtain the representation for each node. Although the topological structure information is investigated, these algorithms ignore the attributes of nodes.

Some graph methods utilize the attributes of nodes based on text and statistical information \cite{ref22,ref23,ref24}. For example, a Graph Convolutional Network (GCN) {method} leverages spectral graph convolutions to extract feature information \cite{ref22}.
A Graph Attention Network (GAT) method specifies the contribution of different neighbors by an attention mechanism \cite{ref23}. GraphSAGE can sample and aggregate neighbor nodes to update the embedding of nodes flexibly\cite{ref24}. A Relational Graph Convolutional Network (RGCN) method can model relational data \cite{ref25}.

Recent studies have handled large-scale graphs and overcome the problem of computational complexity \cite{ref26,ref27}. Considering that a heterogeneous graph contains a large number of types of nodes and edge information, as well as their changes over time, researchers have proposed heterogeneous GNNs \cite{ref29}, and dynamic GNNs \cite{ref32}. In addition, the interpretability and structural optimization of GNNs are also studied \cite{ref33}. 

However, applying the above GNNs to our transaction fraud detection problem fails to utilize all possible features to construct a graph. The graph built in this way may lack much vital information. A competitive graph neural network (CGNN) method \cite{Zhang} utilizes a heterogeneous graph to model normal and fraudulent behaviors in eCommerce. By a 3D convolutional mechanism, a spatial-temporal attention-based graph network (STAGN) method \cite{Cheng} can detect fraudulent transactions from a location-based transaction graph. MAFI utilizes aggregator-level and relation-level attention to learn neighborhood information and different relation \cite{ref59}. LGM-GNN uses local and global information to learn more discriminative representations for prediction tasks \cite{ref60}. Although these methods have tried to learn more discriminative representations, they ignore that excessive aggregation makes indistinguishable representations of nodes in different classes. It results in the over-smoothing phenomenon in GNNs \cite{ref16}. In the applications of transaction fraud detection, fraudsters often disguise cardholders' behaviors by various means. Thereby, there are some edges among fraud nodes and legitimate ones \cite{ref11, ref17}. It exacerbates the influence of the over-smoothing phenomenon. 

\begin{table}[!t]
	\centering
	\caption{NOTATIONS AND EXPLANATIONS.}
	\label{tab:1}       
 \resizebox{\linewidth}{!}{  
\begin{tabular}{l|l}
\toprule
 \textbf{Notation}&\textbf{Description} \\
 \noalign{\smallskip}\hline\noalign{\smallskip}
 $r$     & Transaction record \\
 $m$     & The number of attributes in a transaction record \\
 $C$     & Set of label classes   \\
 $y$     & Label of a transaction record \\
 $\hat{y}$     & The predicted label of a transaction record \\
 $\mathcal{R}$     &  Transaction records set \\
 $\mathcal{G}$     & Graph \\
 $\mathcal{V}$     & Node set \\
 $\mathcal{E}$     & Edge set \\
 $P$     & Logic propositions set \\

 $\mathcal{N}_v$ & Set of neighbors for node $v$ \\
 $\mathbb{A}^k$ & The aggregator function at the $k$-th layer \\ 
 $\hat{z}$ & Neighborhood sample size \\
 $P_{c}(v)$ & \makecell*[l] {The probability of the predicted label class of\\
                            node v is $c$} \\
 $p_{i}$ &  \makecell*[l] {The $i$-th logic proposition based on\\
                      transaction attributes} \\
 $\alpha_{v,v'}^k$ & \makecell*[l]{The attention score between nodes $v$ and $v'$ \\                   at the $k$-th layer}\\
 $\widetilde{\delta} t_{v,v'}$ &  \makecell*[l] {Normalised time interval between two nodes\\
                       $v$ and $v'$} \\

  $h_{v}^{k}$ &  \makecell*[l]{The hidden representation of node $v$ outputted\\ 
                  by the $k$-th layer} \\
 $g_{v}^{k}$ & \makecell*[l]{ The degree gate of node $v$ at the $k$-th layer
                 } \\
\bottomrule
\end{tabular}}
\end{table}

Facing the camouflage issue of fraudsters, the CAmouflage-REsistant GNN (CARE-GNN) method \cite{ref11} defines a label-aware similarity measure using the $l_1$-distance to select neighbors. However, it only focuses on the similarity of labels. $l_1$-distance loses its effectiveness in high-dimensional space and fails to describe the similarity among complex behaviors. In our previous work \cite{ref53}, we measure the distance using cosine similarity, which makes up for the shortcoming of $l_1$-distance. The TG constructed according to transaction data in \cite{ref53} can focus on the relationship among dynamic transaction behaviors, static transaction attributes, and transactions themselves. However, it ignores that fraudsters avoid trades with others to cover up their behaviors, which results in the lack of links among fraudulent nodes. Meanwhile, it fails to tackle the issue of over-smoothing. In this work, we utilize cosine similarity and edge weight to remedy the mentioned flaws and focus on estimating whether neighborhood aggregation is harmful to solving the over-smoothing issue.

\section{Proposed Approach}
This section first describes the preliminaries in the field of transaction fraud detection. After that, ASA-GNN is described in detail. Important notations are listed in Table~1.
\subsection{Preliminary}
\begin{myDef}\emph{Transaction Record.}
	A transaction record $r$ consists of $l$ attributes and a label $y \in \{0, 1\}$.
\end{myDef}

\begin{myDef}\emph{Transaction Graph (TG).}
    A transaction graph is a weighted multigraph $\mathcal{G}=(\mathcal{V}, \mathcal{R}, \mathcal{P},\mathcal{W}eight, \mathcal{E})$, where
    
    1. $\mathcal{V}$ is the set of $|\mathcal{R}|$ nodes and each node $v$ denotes a record $r \in \mathcal{R}$; 
    
    2. $\mathcal{P}$ is the set of $m$ logic propositions that are assertions with respect to the attributes of transaction records and $m\geq 1$;
    
    3. $\mathcal{W}eight: \mathcal{P}\rightarrow \mathbb{N}$ is a weight function; and
    
    4. $\mathcal{E}=\bigcup_{i=1}^{m}\{(a,b)_w^{p_i}|a\in \mathcal{V}\wedge b\in \mathcal{V}\wedge a\neq b\wedge p_i(a,b)=True\wedge w=\mathcal{W}eight(p_i)\}$; 
\end{myDef}

The logic propositions are based on expert rules such that TG ensures the effectiveness of features and reflects the dynamic changes in cardholders' behaviors. In \cite{ref53}, we have defined TG. In comparison with \cite{ref53},
the biggest contributions of this paper lie in the adaptive sampling and aggregation methods, which allow us to utilize GNN to learn the discriminative representations. 

Assume that the underlying graph of a GNN is $\mathcal{G}=\{\mathcal{V},\mathcal{E}\}$. GNNs aim to learn the embedding representations for all nodes and a mapping function such that the predicted results can be obtained. In a general GNN framework \cite{ref15,ref24}, the update function for a single layer is as follows:
\begin{equation}
h_{v}^{k}=\sigma (\mathcal{W}^{k} (h_{v}^{k-1} \oplus \mathbb{A}^{k} (\{ h_{v'}^{k-1}: v' \in  \mathcal{N}_v   \}) )),
\end{equation}
where $h_{v}^{k}$, $\sigma$ and $\mathcal{W}^{k}$ represent the embedding representation of node $v$, activation function, and shared parameter matrix at $k$-th layer respectively.
Given a node $v$, $\mathcal{N}_v$ represents the set of its neighbor nodes, $\mathbb{A}^{k}$ denotes an aggregator function at the $k$-th layer which can aggregate the rich information from $\mathcal{N}_v$ and $\oplus$ is the operator that combines the embedding of $v$ and its neighbor nodes. 

General aggregator functions include a mean aggregator and a pooling one. The mean aggregator is defined as:
\begin{equation}
\mathbb{A}^{k}=  \frac{1}{||\mathcal{N}_v||} \sum_{\substack{v' \in \mathcal{N}_v}}h_{v'}^{k-1}.
\end{equation}

After the $K$ layers' learning, we utilize a classification layer to predict the samples' labels:
\begin{equation}
\hat{y_v}=Softmax(\mathcal{W}^{K}h_{v}^{K}).
\end{equation}
	
In the field of fraud detection, we first construct a TG by transaction records $\mathcal{R}=\{r_1,r_2,...,r_{|\mathcal{R}|}\}$, and then we train the GNN to get the nodes' embedding representations at the last layer and apply a classification layer to predict whether the transaction is fraudulent.

\begin{figure*}
	\centering
	\includegraphics[scale=0.55]{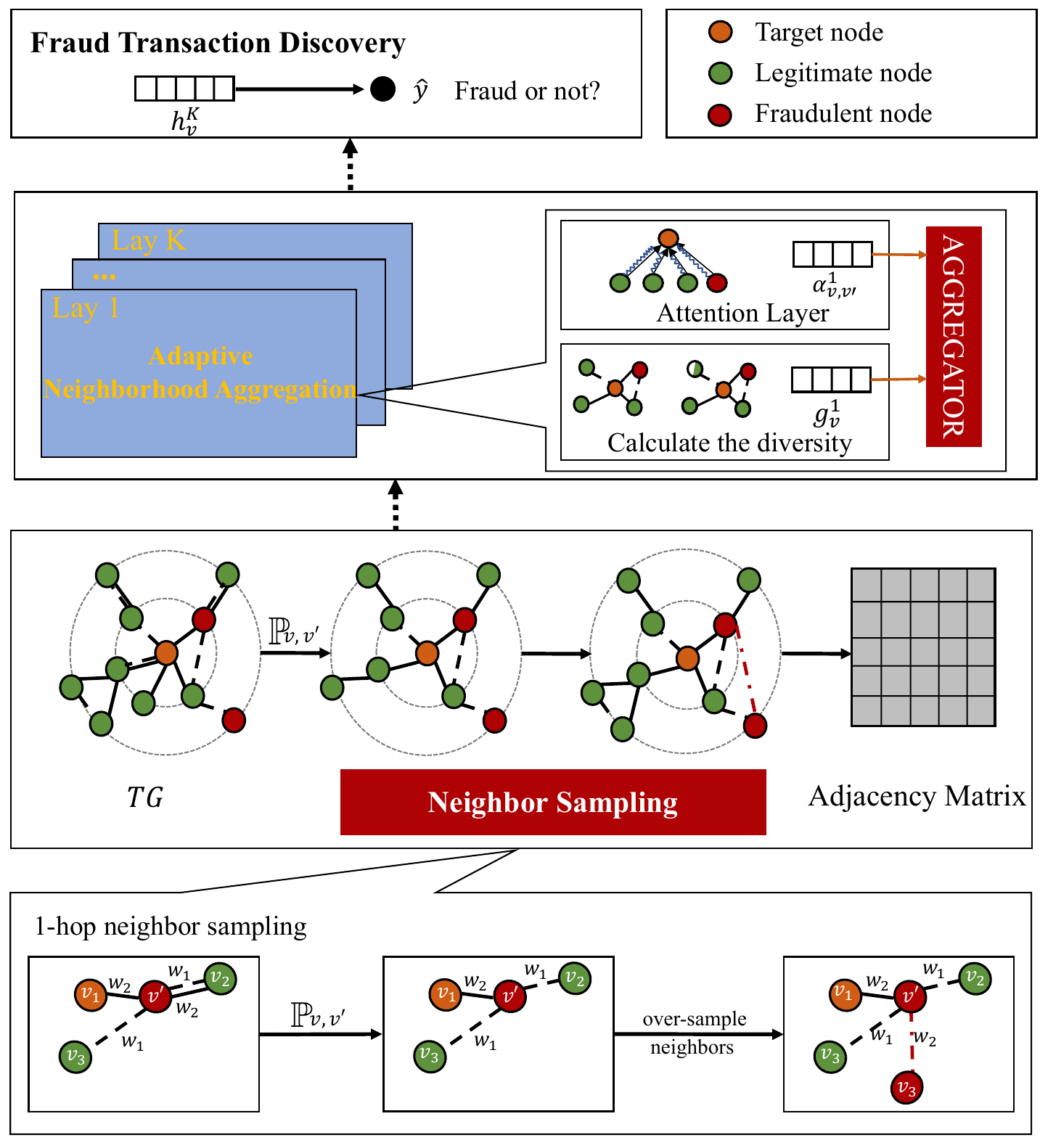}
	\caption{The overview of ASA-GNN: 1) Neighbor sampling at the node level, in which Top-$\hat{z}$ neighbors are sampled filter noise information of each node, and then over-sampling neighbors for fraudulent nodes. 2) Calculating the attention score and aggregation degree to learn representations. 3) Estimating the probability of a transaction being predicted as fraudulent at the detection layer.}
	\label{fig:1.2}       
\end{figure*}

\subsection{ASA-GNN}
The framework of ASA-GNN is illustrated in Fig. 4. Its main components are neighbor sampling and neighborhood aggregation.
In the neighbor sampling stage, to filter noisy neighbors and retain structural information, we define a novel neighbor sampling policy based on cosine similarity and edge weight. In the neighborhood aggregation process, the contributions of different neighbors are specified by an attention mechanism.
After that, we apply a diversity metric to ensure the aggregation degree. Finally, a softmax function calculates the probability of a transaction to predict whether it is fraud. All details are described as follows.

\subsubsection{Neighbor Sampling Strategy Based on Cosine Similarity and Edge Weight}
Simply put, GNNs leverage the information of neighbors to learn more discriminative representations. Existing studies, such as GraphSAGE \cite{ref24}, adopt a random sampling policy under a homophily assumption. However, they ignore the quality of the information from neighbor nodes. Some useless neighbor nodes around a target node result in indistinguishable representations of nodes in different classes. In addition, similar neighbor nodes may provide rich information. Therefore, selecting valid neighbors is necessary before aggregation.

The distance between two nodes and the weight of the edge connecting them are considered to make a novel neighbor sampling policy to deal with this problem. Given a node $v \in \mathcal{V}$ and its neighbor $v' \in \mathcal{V}$, we utilize cosine similarity to compute their distance, which is usually used to analyze user behaviors in practical applications. We calculate the distance between $v$ and $v'$, i.e.,

\begin{equation} \overleftrightarrow{v,v'}= exp(r_{v'} \cdot r_v),\end{equation}
where $r_{v'}$ and $r_v'$ are the normalized attribute vectors of nodes $v$ and $v'$. We utilize the exponential function to ensure non-negative similarity.

Note that there may be multiple edges between two nodes in a TG, and the weight of each edge is assigned. The most significant weight of the edge between $v$ and $v'$ is computed as follows:
\begin{equation} 
w_{v,v'}={max\{\mu_i \cdot Weight(p_i)\}}_{i \in \{1,\cdots,m\}},
\end{equation}
where
\begin{equation}
\mu_i=\left\{
\begin{array}
{ll} 1 &~~\mbox{$p(v, v')=True$} \\
0 &~~~~\mbox{otherwise}.
\end{array}\right.
\end{equation}

Finally, given a node $v \in \mathcal{V}$, the probability of its neighbor $v'$ being selected is defined as

\begin{equation}
\mathbb{P}_{v,v'}= \frac{w_{v,v'} \cdot \overleftrightarrow{v,v'}}{\sum_{\substack{v' \in \mathcal{N}_v, v \neq v'}}w_{v,v'} \cdot \overleftrightarrow{v,v'}},
\end{equation}
where $\mathcal{N}_v$ denotes the set of neighbor nodes of $v$. We perform Top-$\hat{z}$ neighbor sampling to filter noise information from useless neighbor nodes.

After the above neighbor sampling, $\mathcal{N}_v^{'}$ contains the selected neighbor nodes of $v$. However, fraudulent nodes still need neighbors to enrich information. We should find nodes with the same behavior pattern for them. For this purpose, we over-sample neighbors for fraudulent node $v$ as follows:
\begin{equation}
\Bar{\mathcal{N}_v^{f}}
=\{v'\in \mathcal{V}| v' \not\in \mathcal{N'}_v \wedge c_v'=1 \wedge \overleftrightarrow{v,v'}<d_f\},
\end{equation}
where $\overleftrightarrow{v,v'}$ is the distance between nodes $v$ and $v'$ calculated by Eq. (4). Therefore,if $v$ is fraudulent, the set of its neighbors can be defined as follows:
\begin{equation}
\mathcal{N}_v
=\{
\mathcal{N'}_v \cup \Bar{\mathcal{N}_v^{f}}
\}.
\end{equation}
If $v$ is legitimate, the set of its neighbors $\mathcal{N}_v$ can be updated by $\mathcal{N'}_v$.

\subsubsection{Attention Mechanism}
After the neighbor sampling process, $\mathcal{N}_v$ contains the selected neighbor nodes of $v$. Then an aggregator function can generate the embedding representations of $v$ at each layer.
Given a node $v$, $h_{v}^{k}$ denotes the representation at the $k$-th layer where $v \in \mathcal{V}$ and $k=1,2,...,K$. Then it aggregates the information from $\mathcal{N}_v$, which is the set of selected neighbor nodes, i.e.,
\begin{equation}
h_{\mathcal{N}_v}^k= \alpha_{v,v'}^k \cdot \mathbb{A}^k(h_{v'}^{k-1}, \forall v' \in \mathcal{N}_v),
\end{equation}
\begin{equation}
\alpha_{v,v'}^k=\frac{exp(LeakyReLU(e_v^{v'}))}
                    {\sum_{i \in \mathcal{N}_v}{exp(LeakyReLU(e_v^i))}},
\end{equation}

\begin{equation}
e_v^{v'}=f(\mathcal{W}^{k}h_{v}^{k} || \mathcal{W}^{k}h_{v'}^{k}),
\end{equation}
where $\alpha_{v,v'}^k$ denotes the attention score of $v$ and $v'$ at the $k$-th layer, $\mathbb{A}^k$ is an aggregator function at the $k$-th layer, $LeakyReLU$ is an activation function, $f$ is a function mapping the high-dimensional feature to a real number and $\mathcal{W}^{k}$ is a shared parameter matrix.

Generally, the interaction between two transaction records within a short interval is more important. Therefore, given a node $v$ and its neighbor $v'$, the attention score between them at the $k$-th layer is adjusted by the normalised time interval $\{\widetilde{\delta} t_{v,v'}, \forall v' \in \mathcal{N}_v\}$, i.e.,

\begin{equation} 
\alpha_{v, v'}^k = {\widetilde{\delta} t_{v,v'} \cdot \frac{exp(LeakyReLU(e_v^{v'}))}
                    {\sum_{i \in \mathcal{N}_v}{exp(LeakyReLU(e_v^i))}}}.
\end{equation}

\begin{figure*}
	\centering
	\includegraphics[scale=0.9]{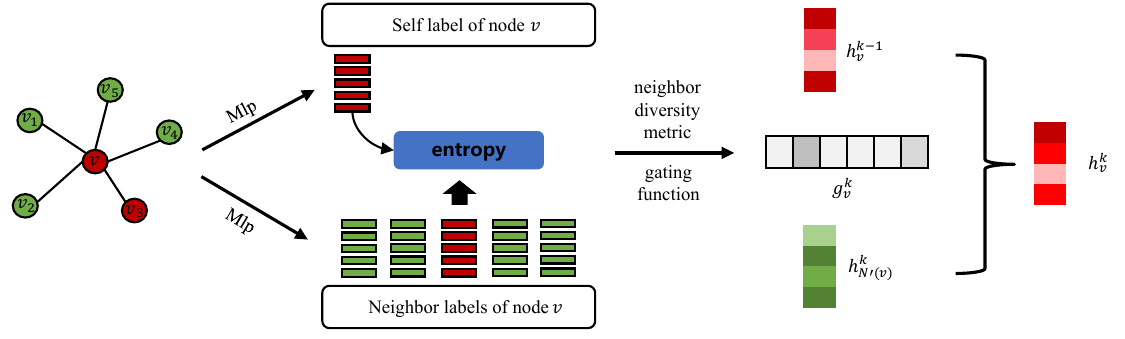}
	\caption{The process of adaptive neighborhood aggregation.}
	\label{fig:1.3}       
\end{figure*}

\subsubsection{Adaptive Neighborhood Aggregation}

Over-smoothing is a common problem in GNN methods. Existing methods assume that the introduction of noise in the aggregation process causes the problem. Specifically, the information from neighbor nodes in the same class makes the representations maintain compactness within the class, which reflects the advantages of GNN. Interactions between a target node and its neighbors in different classes may result in indistinguishable representations. Although neighbor sampling can help us filter some noisy nodes, the camouflage issue of fraudsters brings another challenge. In applications of transaction fraud detection, fraudsters often disguise the cardholders’ behaviors by various means so that there exist edges connecting fraud nodes and legitimate ones in a TG. It exacerbates the effect of the over-smoothing issue. Therefore, when a node has a neighbor in a different class, we should consider that the neighbor may be noisy. 
We introduce a neighbor diversity metric $\mathcal{D}$ by computing the entropy among neighbors of a target node, i.e., 
\begin{equation}
\begin{aligned}
\mathcal{D}(v)=-\sum_{c \in C}P_{c}(v)log(P_{c}(v)),
\end{aligned}
\end{equation}
\begin{equation}
\begin{aligned}
P_{c}(v)=\frac{|{v' \in \mathcal{N}_v|y_{v'} \in c}|}{|\mathcal{N}_v|},
\end{aligned}
\end{equation}

\noindent where $C$ represents the set of label classes, including legitimate and fraudulent ones. $y_{v'}$ is the label of $v'$. The greater the value of $\mathcal{D}(v)$ is, the more diverse the neighbors of $v$ are.

Considering that each node has a $\mathcal{D}$, we use a gating function to control the aggregation degree, i.e.,
\begin{equation}
\begin{aligned}
g_{v}^{k}=\sigma(-Norm(\mathcal{D}(v))), \forall v \in \mathcal{V},
\end{aligned}
\end{equation}
\noindent where $Norm$ is the batch normalization for all nodes in a TG. The range of $g_{v}^{k}$ is (0, 1). When plenty of noisy neighbors are connected to target node $v$, it is very small and close to 0. 

Using the gating function, we allow each node to have its neighborhood aggregation degree. To better understand our adaptive neighborhood aggregation process, the interaction operations of a target node and its neighbors are described, as shown in Fig. 5. The update function for a single layer is as follows:
\begin{equation}
\begin{aligned}
h_{v}^{k}=\sigma(\mathcal{W}^{k} \cdot concat (h_v^{k-1},g_{v}^{k}h_{\mathcal{N}_v}^{k})).
\end{aligned}
\end{equation}

\subsubsection{Detection Layer}
For the target node $v$,  $h_{v}^{K}$  is the final representation outputted by the $K$-th layer. 
After that, a softmax function can be applied to estimate the probability of a transaction being fraudulent. The loss function is computed as follows:
\begin{equation}
\mathcal{L}=  \sum_{i}^{|\mathcal{R}|}{-[y_i \cdot log(\hat{y_i})+(1-y_i) \cdot log(1-\hat{y_i})]},
\end{equation}
where $y_i$ and $\hat{y_i}$ are the labels of the $i$-th transaction record, and the possibility that the sample is predicted to be fraudulent, respectively, and $|\mathcal{R}|$ represents the number of transactions.

\begin{algorithm}[ht]
\footnotesize
\caption{ASA-GNN Approach}
\LinesNumbered 
\KwIn{TG $\mathcal{G}=(\mathcal{V},      \mathcal{R}, \mathcal{P},\mathcal{W}eight, \mathcal{E})$,\\
\quad \quad \quad number of layers $K$, \\
\quad \quad \quad neighbourhood sample size $\hat{z}$, \\
\quad \quad \quad edge weight $\{Weight_1,...,Weight_n\}$,\\ 
\quad \quad \quad non-linear activation function $\sigma$.}
\KwOut{embedding representation $h_v^K$ for each node}
$h_v^0 \leftarrow r_v, \forall v \in \mathcal{V}$; // Initialization   \\ 
\For{each layer $k=1,2,\cdots,K$}{

\For{each $i = 1,2, \cdots ,\hat{z}_k$}
{
// Neighbor sampling \\
$\mathcal{N}_v^i \leftarrow$ select neighbors~from $\mathcal{N}_v$ 
\quad according to Eq.(7)\;
\If{$c_v=1$}{
        over-sample neighbors according to Eq.(8)\;
    }
}

\For{each node $v \in \mathcal{V}$}
{
// Aggregation \\

$\alpha_{u,v}^k \leftarrow$ Eq.(11) \;
$h_{N(v)}^k \leftarrow$ Eq.(8) \;
$g_{v}^k \leftarrow$ Eq.(14) \;
$h_{v}^k \leftarrow$ Eq.(15) \;
}
$h_v^k \leftarrow h_v^k/{||h_v^k||}_2, \forall v \in \mathcal{V}$ \;
    
}

\end{algorithm}

The training process of ASA-GNN is illustrated in Algorithm 1. Given a multigraph $\mathcal{G}$, we first compute the selection probability and then sample $\hat{z}$ neighbors for each node. Then we can compute attention score $\alpha_{u,v}^k$ and aggregation degree $g_{v}^k$. Finally, the representation for each node at the $k$-th layer can be obtained by utilizing an aggregator.


\section{Experiments}
Based on three real-world financial datasets, we conduct the following experiments to show the advantages of ASA-GNN.

\subsection{Datasets and Graph Construction}
\subsubsection{Datasets}
We conduct experiments on one private dataset and two public datasets to demonstrate that ASA-GNN achieves significant improvements over both classic and state-of-the-art models for transaction fraud detection tasks.

The private dataset, PR01, consists of 5.133.5 million transactions from a financial company in Chinathat took place during the second quarter of 2017. Transactions are labeled by professional investigators of a Chinese bank, with 1 representing fraudulent transactions and 0 representing legitimate ones. In data preprocessing, we first utilize the down-sampling of legitimate transactions to solve the imbalanced problem. Then, we apply one-hot coding and min-max normalisation to handle the discrete and continuous values, respectively. Since CARE-GNN requires a lot of computing resources, we take the latest 10000 transaction records as a small dataset (PR02) to facilitate the test.

The TC dataset\footnote{https://challenge.datacastle.cn/v3/} contains 160,764 transaction records collected by Orange Finance Company, including 44,982 fraudulent transactions and 115,782 legitimate transactions. According to the trade time of these transaction records, the training and test sets are divided. Transaction records of one week form the training set and transaction records of the next week form the test set. In this way, the TC dataset is split into TC12, TC23, and TC34. We perform the same data processing as for the PR01 and PR02 datasets.

The XF dataset is a subset extracted from iFLYTEK\footnote{http://challenge.xfyun.cn/2019/gamedetail?type=detail/mobileAD} which have 20000 records. It contains five types of information, including basic data, media information, time, IP information, and device information. The XF dataset is balanced. Therefore, We only perform the same data processing as the datasets PR01 and PR02 to handle the discrete and continuous values.

\begin{table}[!t]
	\caption{EXAMPLES OF TRANSACTION RECORDS.\label{tab:2}}
	\centering
 \resizebox{\linewidth}{!}{  
	\begin{tabular}{|c|c|c|c|c|}
		\hline
		 &  Trade\_time & Trade\_ip & Trade\_mac & Amount\\ 
		\hline
		$r_1$ & 20200618 09:15 & 10.168.31.193 & 15-40-40-50-00-00 & 800 \\
		\hline
		$r_2$ & 20200618 09:20 & 10.168.31.193 & 11-40-50-50-00-00 & 300 \\
		\hline
		$r_3$ & 20200618 09:21 & 10.168.31.193 & 15-40-40-50-00-00 & 200 \\
		\hline
		$r_4$ & 20200618 09:25 & 10.168.31.193 & 12-40-50-50-00-00 & 150 \\
		\hline
		$r_5$ & 20200618 09:35 & 15.168.31.193 & 11-40-50-50-00-00 & 650 \\
		\hline
		$r_6$ & 20200618 10:15 & 15.168.31.193 & 12-40-00-50-00-00 & 930 \\
		\hline
	\end{tabular}
 }
\end{table}



\subsubsection{Graph Construction}

To construct the TG, the transactions are regarded as nodes. Then, we utilize some logic propositions to design the edges. Generally, fraudsters often have two characteristics: device aggregation and temporal aggregation. Device aggregation means that fraudsters are often limited by financial and regulatory constraints and commit fraud on a small number of devices. It differs from legitimate transactions, where cardholders trade on different devices. Temporal aggregation means that fraudsters must complete the fraud activities as quickly as possible since the banks and cardholders may otherwise discover their activities. Therefore, we construct a TG for the private dataset using two logic propositions as follows:

\begin{equation}
p_1(a,b)=\left\{
\begin{array}
{ll} True &~~\mbox{$a.Trade\_ip=b.Trade\_ip~\wedge$} \\ 
&~~\mbox{$|a.Trade\_time-b.Trade\_time|$}\\
&~~\mbox{$\leq 0.5h$}\\
False &~~~~\mbox{otherwise}.
\end{array}\right.
\end{equation}

\begin{equation}
p_2(a,b)=\left\{
\begin{array}
{ll} True &~~\mbox{$a.Trade\_mac=$} \\ 
&~~\mbox{$b.Trade\_mac~\wedge$}\\
&~~\mbox{$|a.Trade\_time-b.Trade\_time|$}\\
&~~\mbox{$\leq 0.5h$}\\
False &~~~~\mbox{otherwise},
\end{array}\right.
\end{equation}
where $Trade\_ip$, $Trade\_time$, and $Trade\_mac$ are the Internet Protocol address, time and Media Access Control address of the transactions, respectively. 

\begin{table*}[ht]
	\centering
	\caption{\centering{PERFORMANCE COMPARISON OF ASA-GNN AND ALL BASELINES ON THE PR01, PR01, XF, TC12, TC23, AND TC34 DATASETS FOR TRANSACTION FRAUD DETECTION TASKS.}}
	\label{tab:3}   
	\scalebox{1.05}{
	\begin{tabular}{l|c|cccccccc}
		\toprule
		Dataset     & Criteria     &  GCN    & GraphSAGE & GAT  &CARE-GNN   &SSA    &RGCN &HAN   & ASA-GNN\\
            \midrule
  		  &                                           $Recall$(\%)   & 65.5  &  67.3 &  86.3  & -   & 86.7  &78.5&86.2& \textbf{88.4}\\
		\textbf{PR01}      &               $F_1$(\%)      &  59.4 &  59.8 &  73.5 &  -  &  71.8 &87.1&75.1& \textbf{90.4}\\
		  &                                           $AUC$(\%)      &  55.7 &   57.6&  69.3  &  -  & 76.4  &73.9&69.6& \textbf{92.2}\\
     \midrule
		  &                                          $Recall$(\%) &  76.1 & 69.9  &   74.7 &  45.8  & 73.0  &70.3&70.1&\textbf{86.1} \\
		 \textbf{PR02}  &    $F_1$(\%)      &  81.1 &  80.7 &  81.5  & 28.7   & 80.4  & 81.1&80.2&\textbf{90.9} \\
		  &                                           $AUC$(\%)      & 75.6 &   81.8&   79.7 &  79.0  &  82.0 &82.4&82.5& \textbf{91.6}\\
		\midrule
		  &                                           $Recall$(\%)   &  68.7 &  65.7 &  68.8  &   50.0 & 65.8  &68.3&67.1& \textbf{73.1}\\
		\textbf{XF}     &                      $F_1$(\%)      &  58.1 &  57.1 &  58.3  &  32.5  & 57.3  &58.5& 52.2& \textbf{59.3} \\
		  &                                            $AUC$(\%)     &   53.3&  52.4 &  48.2  & 48.3   & 52.7  &53.3&52.8& \textbf{57.1}\\
            \midrule
		  &                                            $Recall$(\%) & 57.4  & 55.3  & 58.9    &  62.8  &  58.7 &53.5&56.6& \textbf{73.7}\\
		 \textbf{TC12}     &   $F_1$(\%)      & 67.8 & 63.8 & 66.9   &  70.2  &  69.7 &66.1&61.2& \textbf{77.4}\\
		  &                                            $AUC$(\%)      & 56.9 & 61.4 & 77.1    &  68.8  &  75.8 &65.2&72.9& \textbf{83.5}\\
		\midrule
		  &                                                $Recall$(\%)   & 26.8 & 51.6 & \textbf{66.6} & 62.1&59.1 &52.1&63.5& 66.5\\
		\textbf{TC23}     &   $F_1$(\%)      & 42.3 & 67.4  & 66.6   &   69.6  & 70.4 &67.6&69.1& \textbf{71.1}\\
		  &                                                $AUC$(\%)      & 76.0 & 75.5 & 75.9    &  66.6 &  60.4 &63.7&59.7& \textbf{81.0}\\
		\midrule
		  &                                                $Recall$(\%)   & 47.0 & 50.6 & 62.9    & 66.5    &63.8 &56.7&51.3& \textbf{67.3}\\
		\textbf{TC34}     &     $F_1$(\%)      & 61.6 & 67.2  & 62.8   &  63.7   & 67.9 &72.4&67.8& \textbf{74.6} \\
		  &                                                $AUC$(\%)     & 58.2 & 51.2  & 71.5    &  74.1    &72.8 &54.5&63.0& \textbf{80.3}\\
        
		\bottomrule
	\end{tabular}
	}
\end{table*}

\subsection{Baselines}

To verify the effectiveness of ASA-GNN, the general GNN models and state-of-the-art GNN-based fraud detectors are selected for comparison. The general GNN models includes GCN \cite{ref24}, GraphSAGE \cite{ref24}, GAT \cite{ref23}, RGCN \cite{ref25} and HAN \cite{ref58}. The state-of-the-art GNN-based fraud detectors include CARE-GNN \cite{ref11} and SSA \cite{ref53}.

\begin{itemize}
\item \textbf{GCN} \cite{ref22}: The GCN method leverages spectral graph convolutions to extract feature information.
     \item \textbf{GraphSAGE} \cite{ref24}: GraphSAGE can get a representation for each node using an update function which includes a random sampling policy and neighborhood aggregation process.
	\item \textbf{GAT} \cite{ref23}: The GAT method uses graph attention layers to specify the importance of different neighbors.
    \item \textbf{CARE-GNN} \cite{ref11}: It is a GNN method applied to fraud detection, which improves its aggregation with reinforcement learning to identify the behavior of fraudsters.
	\item \textbf{Similar-sample + attention SAGE (SSA)} \cite{ref53}: The SSA method improves the performance of a model using a sampling strategy and an attention mechanism.
     \item  \textbf{RGCN} \cite{ref25}: RGCN models a relational GNN for link prediction tasks and classification tasks. 
    \item \textbf{HAN} \cite{ref58}: HAN utilizes a hierarchical attention mechanism so that the contributions of different neighbors and meta-paths can be learned. 
\end{itemize}

\subsection{Parameter Settings}
In ASA-GNN, we set $K=3$ as the number of layers, $(20, 20, 20)$ as the neighborhood sample size, $32$ as the hidden size, $0.001$ as the learning rate, $Adam$ as the optimizer and $256$ as the batch size for our PR01 and PR02 datasets. We set $K=3$ as the number of layers, $(30, 50, 50)$ as the neighborhood sample size, $16$ as the hidden size, $0.01$ as the learning rate, $Adam$ as the optimizer and $128$ as the batch size for the XF, TC12, TC23, and TC34 datasets.
For all baseline algorithms , their parameters are the same as those in the corresponding papers\cite{ref22,ref24,ref23,ref11,ref53,ref25,ref58}.

\subsection{Evaluation Criteria}
To measure the performance, we choose $Recall$, $F_1$, and Area Under the Curve of ROC ($AUC$) as criteria. 
$Recall$ represents the ratio of the identified fraudulent transaction records to all fraudulent ones. $F_1$ is a common evaluation criteria in binary classification problems \cite{ref60}. $AUC$ is usually computed to evaluate a model on an imbalanced dataset. $Recall$ and $F_1$ are calculated as follows:

\begin{equation}
Recall= \frac{T_P}{T_P+T_N},
\end{equation}
\begin{equation}
Precision= \frac{T_P}{T_P+F_P},
\end{equation}
\begin{equation}
F_1= \frac{2 \times Recall \times Precision}{Recall+Precision},
\end{equation}
where $T_P$, $T_N$, and $F_P$ are the numbers of true positive transaction records, true negative transaction records, and false positive transaction records, respectively.

$AUC$ is calculated as follows:
\begin{equation}
AUC= \frac{\sum_{r \in \mathcal{R^{+}}}rank_{r}- {\frac{|\mathcal{R^{+}}| \times (|\mathcal{R^{+}}|+1)}{2}} }{  |\mathcal{R^{+}}| \times |\mathcal{R^{-}}|  },
\end{equation}
where $\mathcal{R^{+}}$ and $\mathcal{R^{-}}$ are the fraudulent and legitimate class sets and $rank_{r}$ is the rank of $r$ by the predicted score.


\subsection{Performance Comparison}
The performance of ASA-GNN and all baselines are presented in Table.~3.
The ROC curves of ASA-GNN and all baselines are shown in Fig. 8. We have the following observations and analysis results: 
\begin{itemize}
	\item The proposed ASA-GNN achieves significant improvements over all baselines on the PR01, PR02, XF, TC12, and TC34 datasets. ASA-GNN improves significantly by 6.8\% and 6.7\% in terms of $F_1$ and $AUC$ on the TC12 dataset. Therefore, the overall performance demonstrates the superiority of the proposed ASA-GNN.
    \item GCN, GraphSAGE, GAT, RGCN, and HAN are traditional GNNs, neither of which can identify the camouflage behavior of fraudsters. Thus, their performance is worse than ASA-GNN.
	\item GraphSAGE, CARE-GNN, and SSA are all graph algorithms based on node sampling. None of them performs better than ASA-GNN. The reason is that the proposed ASA-GNN filters nodes effectively and supplements the information of minority nodes, i.e., fraud information. In addition, ASA-GNN considers the camouflage behaviors of fraudsters. The performance of GraphSAGE is worse than that of SSA because noise information may be absorbed in the former's sampling process, and the importance of different nodes needs to be considered.
	\item CARE-GNN calculate the $l_1$-distance between nodes. However, it only focuses on the similarity of labels and the $l_1$-distance loses its effectiveness in high-dimensional space. Although it tries its best to solve the camouflage issue of fraudsters, it still performs poorly.
\end{itemize}

\section{Conclusion and Future Work}
In this paper, a novel graph neural network named ASA-GNN is proposed to identify fraudulent transactions. ASA-GNN employs the neighbor sampling strategy to filter noisy nodes and make up for the lack of neighbors of fraudulent nodes. Consequently, it can make full use of attribute and topology information in TGs. Besides, ASA-GNN can address the camouflage issue of fraudsters and alleviate the over-smoothing phenomena, benefiting from our neighbor diversity metric. Extensive experiments on three financial datasets show that the proposed ASA-GNN achieves significant performance improvements over traditional and state-of-the-art methods. Therefore, ASA-GNN can better help banks and financial institutions detect fraudulent transactions and establish trust relationships with customers.

Our plan includes designing an explainer for the detection model produced by ASA-GNN since the lack of explanations may make customers distrust financial institutions \cite{ref100}. Our TG is built based on expert rules, which can provide a feasible way to develop such an explainer. Studying the imbalance issues in transaction graphs and adding temporal modules (TCN/Transformer) to improve the ability to capture temporal features is also interesting.

\bibliographystyle{IEEEtran}

\bibliography{TLWZ-TCSS-2023-6-8}


 





\end{document}